\documentclass[letterpaper, 10 pt, conference]{ieeeconf}

\IEEEoverridecommandlockouts
\usepackage{graphics}
\usepackage{epsfig}
\usepackage{mathptmx}
\usepackage{times}
\usepackage{amsmath}
\usepackage{amssymb}
\usepackage{xcolor}
\usepackage{mathtools}
\usepackage{siunitx}
\usepackage{algorithm}
\usepackage{algpseudocode}
\usepackage{booktabs}
\usepackage{tabularx}
\usepackage[noadjust]{cite}
\usepackage{stfloats}
\usepackage{graphicx}
\usepackage{censor}
\usepackage{caption}
\usepackage{multirow}

\usepackage{fancyhdr}

\fancypagestyle{arxivnotice}{
    \fancyhf{}%
    \fancyhead[C]{%
        \parbox{\textwidth}{%
            \centering
            \large\color{gray}
            This paper has been accepted for publication at the\\
            IEEE/RSJ International Conference on Intelligent Robots and
            Systems (IROS), 2026.\ \copyright\ IEEE
        }%
    }%
}

\title{\LARGE \bf
Learn Structure, Adapt on the Fly: Multi-Scale Residual Learning and
Online Adaptation for Aerial Manipulators
}

\author{
{Samaksh Ujjawal}$^{1}$,
{Naveen Nair}$^{1}$,
{Shivansh Pratap Singh}$^{1}$,
{Rishabh Dev Yadav}$^{2}$,
{Wei Pan}$^{3}$,
{Spandan Roy}$^{1}$%
\thanks{This work is supported partly by ``Edge-AI-GGCNN'' project from
Qualcomm Technologies and partly by the `UASAT' project sponsored by
MeITY, India.}%
\thanks{$^{1}$Robotics Research Center, IIIT Hyderabad, India.
Emails: \texttt{\{samaksh.ujjawal, shivansh.singh\}@research.iiit.ac.in},
\texttt{naveennair2003@gmail.com},
\texttt{spandan.roy@iiit.ac.in}}%
\thanks{$^{2}$Department of Computer Science,
University of Manchester, UK.
Email: \texttt{rishabh.yadav@postgrad.manchester.ac.uk}}%
\thanks{$^{3}$Newcastle University, UK.
Email: \texttt{wei.pan2@newcastle.ac.uk}}%
}

\begin{document}

\maketitle
\thispagestyle{arxivnotice}


\maketitle

\setlength{\belowcaptionskip}{-10pt}

\begin{abstract}
Autonomous Aerial Manipulators (AAMs) are inherently coupled, nonlinear systems that exhibit nonstationary and multiscale residual dynamics, particularly during manipulator reconfiguration and abrupt payload variations. 
Conventional analytical dynamic models rely on fixed parametric structures, while static data-driven model assume stationary dynamics and degrade under configuration changes and payload variations.
Moreover, existing learning architectures do not explicitly factorize cross-variable coupling and multi-scale temporal effects, conflating instantaneous inertial dynamics with long-horizon regime evolution.
We propose a predictive-adaptive framework for real-time residual modeling and compensation in AAMs. The core of this framework is the Factorized Dynamics Transformer (FDT), which treats physical variables as independent tokens. This design enables explicit cross-variable attention while structurally separating short-horizon inertial dependencies from long-horizon aerodynamic effects. To address deployment-time distribution shifts, a Latent Residual Adapter (LRA) performs rapid linear adaptation in the latent space via Recursive Least Squares, preserving the offline nonlinear representation without prohibitive computational overhead. The adapted residual forecast is directly integrated into a residual-compensated adaptive controller. 
Real-world experiments on an aerial manipulator subjected to unseen payloads demonstrate higher prediction fidelity, accelerated disturbance attenuation, and superior closed-loop tracking precision compared to state-of-the-art learning baselines, all while maintaining strict real-time feasibility.
\end{abstract}

\vspace{-1mm}
\section{Introduction}
\label{sec:introduction}

\begin{figure}[h]
\centering
\includegraphics[width=0.98\linewidth]{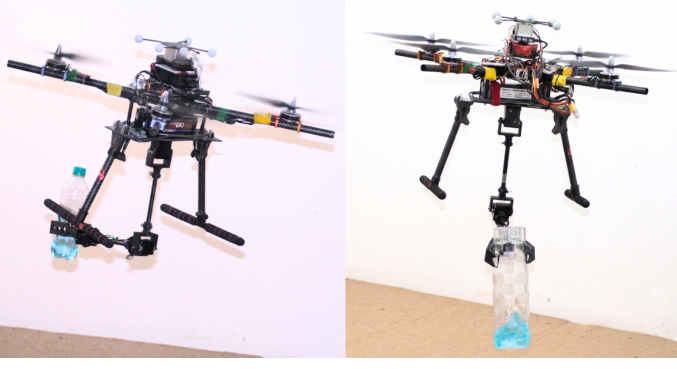}
        \caption{Aerial manipulation platform consisting of a quadrotor and a 2-DOF robotic arm transporting different payloads.}
    \label{fig:teaser}
\end{figure}

Autonomous Aerial Manipulators (AAMs) integrate a quadrotor or multirotor drone with a robotic arm. This setup boosts the drone's dexterity and versatility, enabling tasks from basic payload transport to advanced operations like pick-and-place, contact inspection, grasping, and assembly \cite{ruggiero2018aerial, yadav2025integrated}. 
Nevertheless, this integration introduces fundamentally challenging dynamics that defy classical analytical treatment such as configuration-dependent inertia variations as the arm moves or engages the environment (e.g., during grasping or dropping,  see Fig. \ref{fig:teaser}), nonlinear coupling forces between the drone and arm, and aerodynamic phenomena, such as rotor downwash and manipulator-induced flow interference {\cite{yadav2024modular, ollero2022past}}. These forces collectively dominate system behavior, making it diverge from assumed mathematical model and giving rise to \emph{residual dynamics}.

Conventional AAM control methods \cite{ollero2022past, meng2020survey, sharma2025impedance} rely on analytical knowledge of the underlying system model. Unfortunately, it is difficult, if at all possible, to accurately represent configuration-dependent inertial coupling forces analytically \cite{yadav2024modular}.
Therefore, these traditional control frameworks struggle to compensate for residual dynamics. Data-driven methods \cite{saviolo2023learning, jiahao2023online} have recently found footprints in robotics literature to tackle the issues that arise from model-based methods. However, these existing methods are not directly suitable for the residual dynamics in an AAM system. In the following, we steer our discussion in this direction. 

\vspace{-1mm}

\subsection{Related Works}

Offline data-driven models, including Deep Neural Networks (DNNs)~\cite{shi2019neural}, Gaussian Processes (GPs)~\cite{torrente2021data,cao2024computation}, physics-informed Temporal Convolutional Networks (TCNs)~\cite{saviolo2022physics}, Recurrent Neural Networks (RNNs)~\cite{mohajerin2019multistep}, and diffusion models~\cite{das2025dronediffusion,ujjawal2025aermani}, effectively model quadrotor dynamics. However, static offline models struggle with aerial-manipulator dynamics that vary with manipulator configuration and payload, causing accumulating prediction and control errors without online adaptation~\cite{saviolo2023learning}.

Language and vision models support grasping, placement, and vision-language-based skill selection~\cite{song2025soranav, singh2026aerograb,mishra2026aeroplace,mishra2025aermani}, but generally neglect accurate dynamics. Existing approaches address nonstationarity through controller adaptation~\cite{o2022neural} or online model updates~\cite{saviolo2023active,jiahao2023online,gahlawat2020l1,lew2022safe,richards2021adaptive,yadav2026physics,yadav2025arcade,yadav2026learning}. Last-layer gradient updates~\cite{saviolo2023active} require optimizer tuning and are data-inefficient, whereas full-network retraining with exponential moving averages~\cite{jiahao2023online} is computationally expensive. Meta-learning~\cite{o2022neural,lew2022safe} interpolates among pre-trained models, limiting generalization to unseen dynamics. GP regression~\cite{gahlawat2020l1}, Bayesian last-layer adaptation~\cite{lew2022safe}, and adaptive control~\cite{richards2021adaptive} enable online updates but do not explicitly separate cross-variable coupling from multi-scale temporal effects and often assume gradual, data-rich adaptation.

\paragraph{Multiscale Dynamics}
Aerial-manipulator residuals contain fast configuration-dependent inertial and Coriolis effects, and slower aerodynamic and payload-induced transients~\cite{orsag2017dexterous,ollero2022past}. The former depend mainly on instantaneous states, whereas the latter evolve over several control cycles. Modeling both requires short histories ($1$--$5$ samples) and longer windows ($5$--$20$ or more samples), which standard Transformers do not explicitly separate.

\paragraph{Spatial Tokenization}
Conventional Transformers tokenize time steps, although aerial-manipulator residuals are strongly governed by coupling among physical variables. Temporal attention therefore mixes instantaneous cross-variable interactions with long-term regime memory and lacks a physics-aligned inductive bias.

\paragraph{Online Adaptation}
Payload changes and aggressive maneuvers require adaptation from streaming data without retraining the backbone or destabilizing high-frequency control. This motivates separating offline nonlinear representation learning from lightweight online identification.

Accordingly, effective residual modeling requires factorization across variables, temporal scales, and adaptation modes. Our predictive--adaptive framework (Fig.~\ref{fig:pipeline}) provides:

\begin{enumerate}
    \item \textbf{Structure-Aligned Residual Modeling:}
    The Factorized Dynamics Transformer (FDT) treats physical variables as tokens and uses dual streams to separate short-horizon interactions from long-horizon regime memory.

    \item \textbf{Latent Linear Online Identification:}
    The Latent Residual Adapter (LRA) applies a linear latent-space correction updated through Recursive Least Squares, enabling rapid adaptation while preserving the offline nonlinear backbone.

    \item \textbf{Residual-Compensated Adaptive Control:}
    Adapted residual predictions compensate closed-loop modeling errors through an adaptive controller that avoids complete system identification and adapts only one parameter, independent of system complexity.
\end{enumerate}

Experiments on a real aerial manipulator under unseen payloads demonstrate improved prediction accuracy, faster disturbance recovery, and real-time closed-loop feasibility.

\vspace{-1mm}
\section{Methodology}
\label{sec:methodology}

We consider a quadrotor-based AAM composed of a 6-degrees-of-freedom (DOF) floating base and an $n_{arm}$-DOF robotic arm. 
Let $n \triangleq 6 + n_{arm}$ denote the total number of generalized coordinates, and let $\chi \in \mathbb{R}^{n}$ be the full configuration vector with control input $\tau \in \mathbb{R}^{n}$. The system dynamics follow the Euler--Lagrange formulation \cite{yadav2024modular}
\begin{equation}
\label{eq:true_dynamics}
M(\chi)\ddot{\chi} 
+ C(\chi,\dot{\chi})\dot{\chi} 
+ g(\chi) 
+ d(\dot{\chi},t) 
= \tau,
\end{equation}
where $M(\chi) \in \mathbb{R}^{n \times n}$ is the inertia matrix, 
$C(\chi,\dot{\chi})\dot{\chi}$ represents Coriolis and centripetal effects, 
$g(\chi)$ denotes gravity, and 
$d(\dot{\chi},t)$ captures aerodynamic and external disturbances.

To isolate configuration-dependent and nonstationary effects, we introduce a user-defined constant diagonal matrix $\bar{M} \in \mathbb{R}^{n \times n}$ and rewrite \eqref{eq:true_dynamics} as
\begin{equation}
\label{eq:residual_dynamics}
\bar{M}\ddot{\chi} + \mathbf{r}(\chi,\dot{\chi},\ddot{\chi},t) = \tau,
\end{equation}
where the \emph{residual dynamics} are defined as
$
\mathbf{r}
\triangleq
\big(M(\chi)-\bar{M}\big)\ddot{\chi}
+ C(\chi,\dot{\chi})\dot{\chi}
+ g(\chi)
+ d(\dot{\chi},t).$
The residual $\mathbf r$ aggregates inertial coupling, nonlinear interaction forces, aerodynamic transients, and external disturbances into a single term. 
Rather than estimating each physical component separately, we directly learn $\mathbf r$, which is more suitable for aerial manipulators operating under continuously varying posture and payload conditions.
From logged data, the residual can be computed as
$
\mathbf r_t = \tau_t - \bar{M}\ddot{\chi}_t,
$
allowing supervised learning from measured trajectories.

Given a history window of length $T$, we define the trajectory segment
$
\mathcal D_t
=
\{ (\chi_{t-i}, \dot{\chi}_{t-i}, \tau_{t-i}) \}_{i=0}^{T}.
$
Our objective is to learn a parameterized predictor that produces multi-step residual forecasts
$
\hat{\mathbf r}_{t+j|t},
$
representing the $j$-step-ahead prediction conditioned on data available up to time $t$.
The predictor is parameterized by $\theta$ and maps $\mathcal D_t$ to the prediction horizon.
The model parameters $\theta$ are obtained by minimizing the multi-step prediction loss
\begin{equation}
\label{eq:training_loss}
\mathcal L(\theta)
=
\sum_t
\sum_{j=0}^{k}
\left\|
\mathbf r_{t+j}
-
\hat{\mathbf r}_{t+j|t}
\right\|_2^2.
\end{equation}
This multi-step objective reduces error accumulation during rollout and improves the stability of downstream control by encouraging accurate residual forecasts over the prediction horizon.
The residual process $\mathbf r_t$ depends on configuration-dependent inertial coupling and aerodynamic transients, introducing history-dependent effects not captured by a first-order Markov model. 
We adopt a two-stage strategy:

\begin{enumerate}
\item \textbf{Offline Residual Learning:} 
Factorized Dynamics Transformer (FDT) is trained to approximate the mapping from $\mathcal D_t$ to $\{\hat{\mathbf r}_{t+j|t}\}_{j=0}^{k}$ using large-scale trajectory data.

\item \textbf{Online Latent Adaptation:} 
To handle distribution shifts induced by payload or operating-condition changes, a lightweight linear correction layer is adapted online in latent space using Recursive Least Squares (RLS).
\end{enumerate}

This separation  yields a residual predictor that is both high-capacity and computationally efficient for real-time aerial manipulation.

\begin{figure}[t]
\centering
\includegraphics[width=0.98\linewidth]{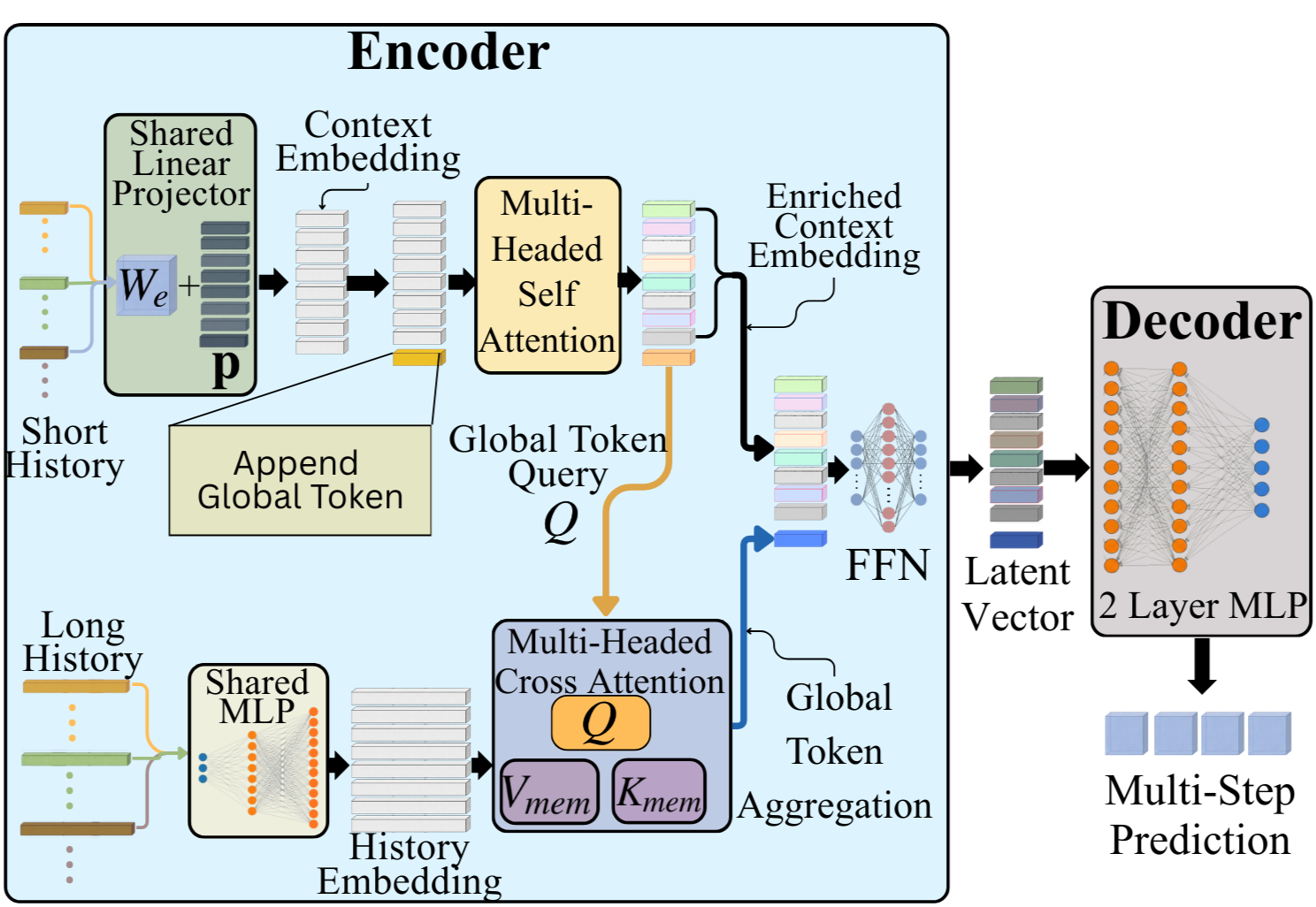}
        \caption{The Factorized Dynamics Transformer (\textbf{FDT}) architecture. A dual-stream encoder separates short-horizon inertial coupling (via Self-Attention) from long-horizon aerodynamic memory (via Cross-Attention). A learnable Global Token aggregates both temporal scales into a unified Latent Vector, which is mapped to multi-step residual predictions by a feed-forward decoder, i.e., a position-wise FFN followed by an MLP.}
    \label{fig:fdt_arch}
\end{figure}

\subsection{Offline Residual Dynamics Learning}
\label{subsec:offline_rdl}

The residual process $\mathbf r_t$ defined in \eqref{eq:residual_dynamics} depends on configuration-dependent inertial coupling and aerodynamic transients, introducing both cross-variable interactions and history-dependent effects. 
Standard Markov models that rely solely on the instantaneous state $(\chi_t,\dot\chi_t)$ are therefore insufficient to capture this structure.
To model these multi-scale dependencies, we propose the \textbf{Factorized Dynamics Transformer (FDT)}  (depicted in Fig. \ref{fig:fdt_arch}), a transformer-based architecture designed to (i) explicitly model inter-variable coupling and (ii) incorporate long-horizon temporal memory without recurrent state compression.

\subsubsection{Variable-Wise Temporal Embedding}
\label{subsubsec:inverted_embedding}

At each time step, the stacked input vector is 
$
\mathbf x_t =
\begin{bmatrix}
\chi_t^\top &
\dot{\chi}_t^\top  &
\tau_t^\top
\end{bmatrix}^\top
\in \mathbb{R}^{3n}.
$
We therefore define the number of scalar input channels as $d_v = 3n$.
Let $z_{i,t}$ denote the $i$-th scalar component of $\mathbf x_t$. 
For each scalar variable $i \in \{1,\dots,d_v\}$, we extract its short-horizon history over $T_s$ steps:
\begin{align}
 \mathbf z_i
=
\begin{bmatrix}
z_{i,t-T_s+1} & \dots & z_{i,t}
\end{bmatrix}^\top
\in \mathbb R^{T_s}.   
\end{align}
Rather than tokenizing along the time dimension, each physical variable is treated as an independent token. 
Each token is embedded into a latent space of dimension $d_{model}$ via
\begin{align}
   \mathbf e_i
=
W_e \mathbf z_i + \mathbf p_i,
\qquad
W_e \in \mathbb R^{d_{model}\times T_s}, 
\end{align}
where $W_e$ is shared across variables and $\mathbf p_i \in \mathbb R^{d_{model}}$ is a learnable embedding encoding the physical identity of variable $i$.
Stacking all embeddings yields the context matrix
\begin{align}
 E_{ctx}
=
\begin{bmatrix}
\mathbf e_1 \\
\vdots \\
\mathbf e_{d_v}
\end{bmatrix}
\in \mathbb R^{d_v \times d_{model}}.  
\end{align}
\subsubsection{Short-Horizon Context Encoding}
\label{subsubsec:context_stream}

To capture instantaneous cross-variable dependencies relevant for control, we append a learnable global token $\mathbf g \in \mathbb R^{d_{model}}$ to the context matrix:
$
[E_{ctx}; \mathbf g] \in \mathbb R^{(d_v+1)\times d_{model}}.
$
Multi-head self-attention (MSA) utilizing 
${n_{heads}}$ parallel attention heads is then applied to the updated context matrix
\begin{align}
 \tilde E
=
\text{MSA}([E_{ctx}; \mathbf g])
+
[E_{ctx}; \mathbf g].   
\end{align}
The output can be partitioned as $
\tilde E = [\tilde E_{ctx}; \tilde{\mathbf g}_{ctx}],
$
where $\tilde{\mathbf g}_{ctx}$ denotes the updated global token from the context branch.
Since self-attention computes pairwise interactions across all tokens, the global token aggregates contextual information from all physical variables, forming a compact latent summary of the joint system state within the short-horizon window. While the attention mechanism does not enforce physical coupling explicitly, the variable-wise tokenization provides an inductive bias that encourages the model to learn dependencies across degrees of freedom.

\subsubsection{Long-Horizon Memory Retrieval}
\label{subsubsec:memory_stream}

Residual dynamics also exhibit slower temporal effects (e.g., wake interaction and 
payload-induced inertia shifts). To incorporate longer memory without quadratic 
self-attention over the entire sequence, we introduce a long-horizon window $T_\ell > T_s$. Simply increasing \(T_s\) would require applying full self-attention over longer temporal contexts, leading to quadratic growth in computational cost and potentially mixing instantaneous dynamics with slowly evolving regime effects. The proposed factorization instead separates these roles: short-horizon attention captures rapid cross-variable interactions, while long-horizon memory provides a compressed representation of slower aerodynamic and payload-induced dynamics.

Let the stacked long-horizon input sequence be
\begin{align}
Z_\ell
=
\begin{bmatrix}
\mathbf{x}_{t-T_\ell+1} & \cdots & \mathbf{x}_t
\end{bmatrix}^\top
\in \mathbb{R}^{T_\ell \times d_v}
\end{align}
We use \emph{Variable-Wise Temporal Embedding} as discussed in Section~\ref{subsubsec:inverted_embedding}, where the $i$-th token
$
\mathbf{z}_i \in \mathbb{R}^{T_\ell}, \qquad i = 1,\ldots,d_v,
$
carries the full long-horizon temporal history of the $i$-th state variable.

A single shared MLP with output embedding dimension $d_{model}$ is applied 
identically to every variable token:
\begin{align}
 \mathbf{e}_i = \mathrm{MLP}(\mathbf{z}_i) \in \mathbb{R}^{d_{model}},
\qquad i = 1,\ldots,d_v.   
\end{align}
Stacking the embeddings yields long-horizon memory matrix
\begin{align}
\mathbf{E}_{hist}
=
\begin{bmatrix}
\mathbf{e}_1 & \cdots & \mathbf{e}_{d_v}
\end{bmatrix}^\top
\in \mathbb{R}^{d_v \times d_{model}}
\end{align}

Memory keys and values are obtained by projecting from $d_{model}$ to $d_k$:
\begin{align}
K_{mem} = \mathbf{E}_{hist} W_K \in \mathbb{R}^{d_v \times d_k}, ~~
V_{mem} = \mathbf{E}_{hist} W_V \in \mathbb{R}^{d_v \times d_k},    
\end{align}
where $W_K, W_V \in \mathbb{R}^{d_{model} \times d_k}$ are learnable projection matrices.

The updated global token $\tilde{\mathbf{g}}_{ctx} \in \mathbb{R}^{d_{model}}$ serves 
as the query:
\begin{align}
 Q = \tilde{\mathbf{g}}_{ctx} W_Q \in \mathbb{R}^{1 \times d_k},
\qquad
W_Q \in \mathbb{R}^{d_{model} \times d_k}.   
\end{align}

Cross-attention over the $d_v$ physical-variable tokens yields
{\small
\begin{align} 
\alpha
= \mathrm{Softmax}\!\left(\frac{Q K_{mem}^\top}{\sqrt{d_k}}\right)
\in \mathbb{R}^{1 \times d_v}, ~~
\mathbf{g}'_{ctx} = \alpha\, V_{mem} \in \mathbb{R}^{d_k},    
\end{align}
}
where the Softmax is applied across the $d_v$ variable tokens, producing an 
attention weight that reflects the relative dynamical relevance of each physical 
channel over the long horizon.

This operation compresses the long-horizon trajectory into a single latent vector 
$\mathbf{g}'_{ctx} \in \mathbb{R}^{d_k}$, which acts as a low-dimensional 
representation of the current dynamic regime and is passed directly to the RLS 
adaptation stage as $\mathbf{g}'_t$.

\subsubsection{Residual Prediction}

The memory-enriched global token $\mathbf g'_{ctx}$, which encapsulates both short-horizon cross-variable interactions and long-horizon regime information, is appended to the context embedding from the MSA, yielding the final encoder output embedding as: 
\begin{align}
 \tilde E_{enc}
=[E_{ctx}; \mathbf g'_{ctx}]   
\end{align}
As shown in the decoder block of Fig. \ref{fig:fdt_arch}, this enriched context  is passed through a position-wise feed-forward network (FFN) with a hidden expansion dimension of ${d_{ff}}$, followed by a multi-layer perceptron to produce multi-step residual predictions:
\begin{align}
 \mathbf h = \text{FFN}(\tilde E_{enc}),
\qquad
\{\hat{\mathbf r}_{t+j|t}\}_{j=0}^{k}
=
\text{MLP}(\mathbf h).   
\end{align}

The parameters of the FDT are trained offline by minimizing the loss in \eqref{eq:training_loss} over a dataset spanning diverse configurations, velocities, and payload conditions.

\begin{figure*}[]
\centering
\vspace{-2mm}
\includegraphics[width=0.95\linewidth]{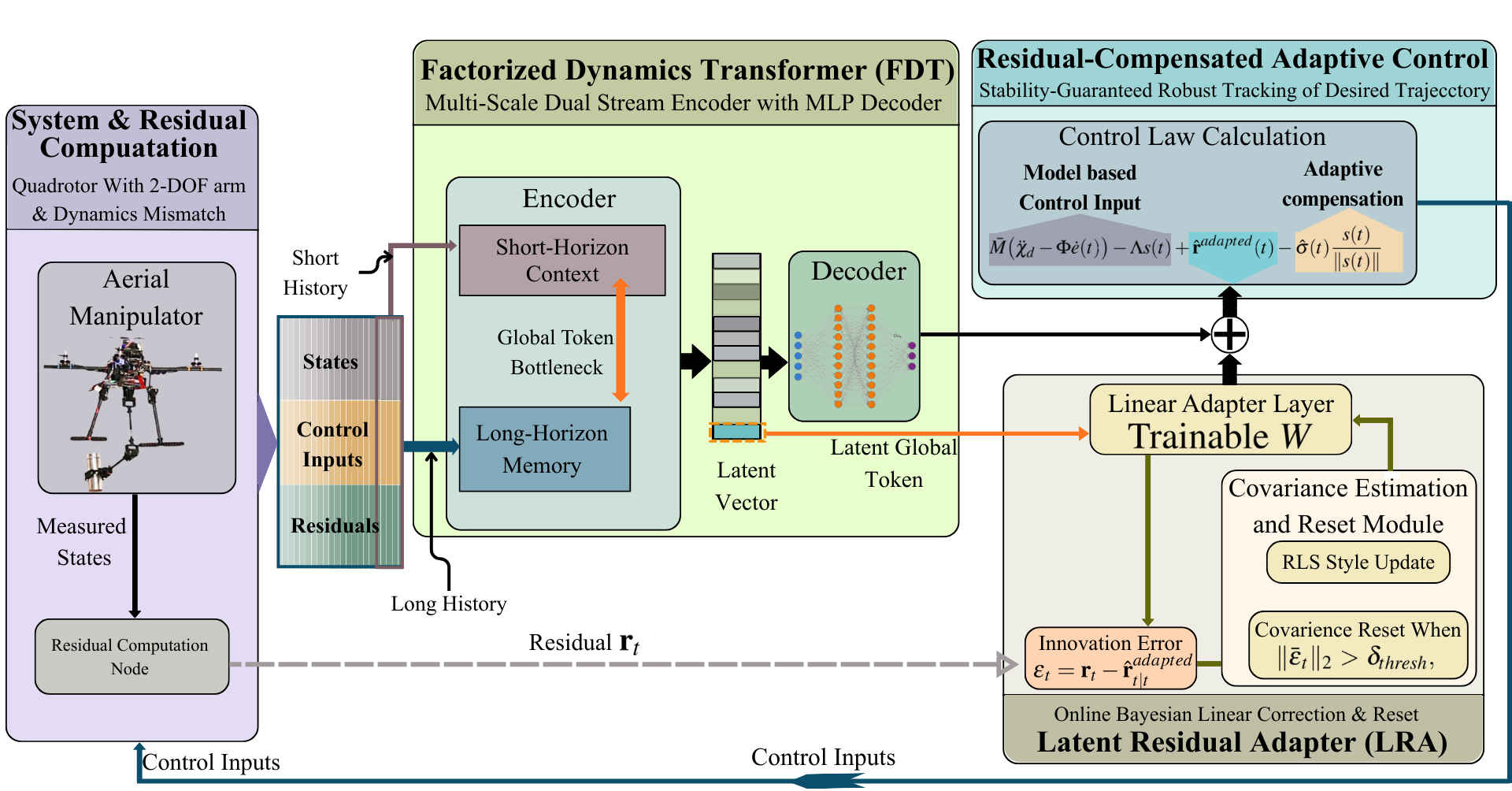}
    \caption{ The framework fuses a physics-aware \textbf{Factorized Dynamics Transformer (FDT)} with an online \textbf{Latent Residual Adapter (LRA}). The FDT (center) utilizes inverted variable embeddings and a global-token bottleneck to efficiently compress long-horizon aerodynamic memory. To handle unmodeled regime shifts (e.g., payload changes), the LRA (bottom-right) performs online Bayesian correction directly on the frozen global token features, injecting a real-time adjustment into the adaptive control law (top-right) to guarantee robust trajectory tracking.    }
    \label{fig:pipeline}
\end{figure*}

\subsection{Online Latent Residual Adaptation}
\label{subsec:online_adaptation}

While the offline FDT captures structured residual dynamics from diverse training data, aerial manipulators frequently experience distribution shifts caused by unseen payloads or abrupt operating-condition changes. 
To enable rapid adaptation without full-network retraining, we introduce a lightweight latent correction mechanism operating on the learned regime representation.

\subsubsection{Additive Latent Residual Model}
\label{subsubsec:_online_lra}

Let $\hat{\mathbf r}^{base}_{t|t} \in \mathbb R^{n}$ denote the one-step residual prediction produced by the frozen FDT. 
From the encoder, we extract the memory-enriched global token 
$\mathbf g'_t \in \mathbb R^{d_k}$, which represents a compact latent embedding of the current dynamic regime.

We assume that deviations from the base model induced by payload or regime shifts are linear-in-parameters with respect to the latent feature:
\begin{align}
\label{lip}
 \mathbf r_t
=
\hat{\mathbf r}^{base}_{t|t}
+
W^\top \mathbf g'_t
+
\boldsymbol{\epsilon}_t,  
\end{align}
where $W \in \mathbb R^{d_k \times n}$ is an adaptation matrix and $\boldsymbol{\epsilon}_t \in \mathbb R^{n}$ denotes residual modeling error.
The adapted residual prediction is therefore
\begin{align} \label{r_adapt}
 \hat{\mathbf r}^{adapted}_{t|t}
=
\hat{\mathbf r}^{base}_{t|t}
+
W_t^\top \mathbf g'_t.   
\end{align}
This structure preserves the expressive nonlinear mapping learned offline while restricting online adaptation to a linear correction in latent space.

\subsubsection{Recursive Least Squares Update}

The adaptation matrix $W_t$ is updated online using Recursive Least Squares (RLS). 
Define the innovation error $
\boldsymbol{\epsilon}_t
=
\mathbf r_t
-
\hat{\mathbf r}^{adapted}_{t|t}.$

Let $\Sigma_t \in \mathbb{R}^{d_k \times d_k}$ denote the covariance matrix and $\lambda \in (0,1]$ a forgetting factor.
The RLS update equations are
\begin{subequations}
\begin{align}
K_t
&=
\frac{\Sigma_{t-1} \mathbf{g}'_t}
{\lambda + \mathbf{g}_t^{\prime\top} \Sigma_{t-1} \mathbf{g}'_t},
\label{eq:rls_gain}
\\
W_t
&=
W_{t-1}
+
K_t \boldsymbol{\epsilon}_t^\top,
\label{eq:rls_weight}
\\
\Sigma_t
&=
\frac{1}{\lambda}
\left(
\Sigma_{t-1}
-
K_t \mathbf{g}_t^{\prime\top} \Sigma_{t-1}
\right).
\label{eq:rls_cov}
\end{align}
\end{subequations}

This update minimizes the exponentially weighted least-squares objective
\begin{align}
 \sum_{\tau=0}^{t}
\lambda^{t-\tau}
\|
\mathbf r_\tau
-
\hat{\mathbf r}^{base}_{\tau|\tau}
-
W^\top \mathbf g'_\tau
\|_2^2.   
\end{align}

\subsubsection{Adaptive Covariance Reset}
\label{reset}

A known limitation of standard Recursive Least Squares (RLS) is covariance collapse ($\Sigma_t \rightarrow 0$), which gradually reduces parameter plasticity and slows adaptation when the system experiences abrupt regime shifts. In practice, this effect can cause the estimator to become overly confident in outdated parameters, preventing rapid adjustment to new dynamics such as sudden payload changes or aerodynamic disturbances.

To mitigate this issue, we introduce an adaptive covariance reset mechanism that restores estimator responsiveness when large prediction errors are observed. 
Let $\bar{\boldsymbol{\epsilon}}_t$ denote the exponential moving average of the innovation error. If $
\|\bar{\boldsymbol{\epsilon}}_t\|_2 > \delta_{thresh}, $
the covariance matrix is reinitialized as
$
\Sigma_t \leftarrow \sigma_0 I,
$
where $\sigma_0 > 0$ controls the adaptation aggressiveness and $I$ is the identity matrix.

This reset temporarily increases the Kalman gain $K_t$, allowing the estimator to rapidly remap the latent feature $\mathbf g'_t$ to the updated residual distribution. In effect, the mechanism restores estimator plasticity and enables fast adaptation under sudden dynamic changes.
While this modification departs from the strict optimality guarantees of classical RLS, it acts as a practical safeguard that improves adaptation speed and robustness in the presence of nonstationary dynamics.

\vspace{-1mm}
\subsection{Residual-Compensated Adaptive Control}
\label{subsec:control}
While the FDT and LRA bring the nominal model closer to reality, a 
estimation error inevitably remains. Given the multi-step residual forecast $\{\hat{\mathbf r}_{t+j|t}\}_{j=0}^{k}$ produced by the proposed learning framework, only the one-step prediction $\hat{\mathbf r}_{t|t}$ is used for feedback compensation. After online adaptation, let $\hat{\mathbf r}^{adapted}_{t|t}$ denote the corrected estimate and $\sigma(t) \triangleq \hat{\mathbf r}^{adapted}(t) - \mathbf r(t)$, bounded over a compact set, represents the residual estimation error.
Denoting the tracking error as
$e(t) \triangleq \chi(t) - \chi_d(t)$, where $\chi_d(t)$ being the desired reference trajectory, we define an error variable $s$ as
\begin{equation}
s(t) = \dot e(t) + \Phi e(t),
\label{eq:sliding_var}
\end{equation}
where $\Phi \in \mathbb R^{n\times n}$ is a user-defined positive definite gain matrix. Multiplying the time derivative of \eqref{eq:sliding_var} by $\bar{M}$ and using \eqref{eq:residual_dynamics} we obtain
\begin{align}
    \bar{M}\dot{s}
      &= \bar{M}(\ddot{\chi} - \ddot{\chi}_d + \Phi\dot{e}) = \tau - \mathbf r - \bar{M}(\ddot{\chi}_d - \Phi\dot{e}).
    \label{eq:open_loop_s}
\end{align}
We propose the following residual-compensated control law:
{\small
\begin{subequations}
\label{eq:adaptive_control_law}
\begin{align}
\tau(t)
&=
\bar{M}\big(\ddot{\chi}_d - \Phi \dot e(t)\big)
-
\Lambda s(t)
+
\hat{\mathbf r}^{adapted}(t)
-
\hat{\sigma}(t)\frac{s(t)}{\|s(t)\|},
\label{eq:control_input}
\\[2mm]
\dot{\hat{\sigma}}(t)
&=
\|s(t)\| - \nu\,\hat{\sigma}(t),
\qquad
\hat{\sigma}(0) > 0,
\label{eq:adaptive_law}
\end{align}
\end{subequations}
}
where $\Lambda \in \mathbb R^{n\times n}$ ($\Lambda \succ 0$) is a control gain matrix, $\nu>0$ is a design constant, and $\hat{\sigma}(t)$ is an adaptive gain designed to counteract bounded residual estimation uncertainties ${\sigma}$. 
Substituting the proposed control input~\eqref{eq:control_input} back into open-loop dynamics~\eqref{eq:open_loop_s} yields the closed-loop error dynamics:
\begin{align}
    \bar{M}\dot{s}
        &= -\Lambda s(t) + \big(\hat{\mathbf r}^{adapted}(t) - \mathbf r(t)\big)
           - \hat{\sigma}(t)\frac{s(t)}{\|s(t)\|} \nonumber\\
        &= -\Lambda s(t) + \sigma(t)
           - \hat{\sigma}(t)\frac{s(t)}{\|s(t)\|}.
    \label{eq:closed_loop_s}
\end{align}
By continuously fine-tuning the latent representation via the online adapter, the magnitude of the estimation error $\sigma(t)$ is drastically minimized. This proactively reduces the burden on the high-frequency adaptive term $\hat{\sigma}(t)\frac{s}{\|s\|}$, mitigating chattering and ensuring tight tracking precision.
Algorithm \ref{algo:inference} and Figure \ref{fig:pipeline} summarize the closed-loop pipeline.

\textit{Remark:} Although a unified end-to-end proof is not derived here, stability can be argued from established results under standard assumptions. The offline residual learner converges in the supervised learning sense given sufficiently rich training data; the Recursive Least Squares update yields bounded parameter estimates under persistent excitation and bounded noise; and the residual-compensated adaptive controller ensures Lyapunov stability under bounded residual estimation error.

\begin{algorithm}[t]
\caption{Real-Time Residual-Compensated Control} \label{algo:inference}
\textbf{Input:} History $\mathcal{D}_t$, frozen FDT, $\bar{M}$, $(\lambda,\sigma_0,\delta_{\text{thresh}})$, $(\Phi,\Lambda,\nu)$ \\
\textbf{Initialize:} $W_0 = 0_{d_k \times n}$, $\Sigma_0 = \sigma_0 I$, $\hat{\sigma}(0) > 0$, $\bar{\boldsymbol{\epsilon}}_0 = 0$
\begin{algorithmic}[1]
\While{system active at time $t$}
    \State Measure $(\chi_t,\dot{\chi}_t,\ddot{\chi}_t)$ and append $\mathcal{D}_t \leftarrow \chi_t,\dot{\chi}_t, \tau_t$
    \State Compute residual $\mathbf r_t$ via \eqref{eq:residual_dynamics}
    \State $(\mathbf g'_t, \hat{\mathbf r}^{base}_{t|t}) \leftarrow$ FDT$(\mathcal{D}_t)$  \hfill // \ref{subsec:offline_rdl}
    \State $\boldsymbol{\epsilon}_t \leftarrow \mathbf r_t - \big(\hat{\mathbf r}^{base}_{t|t} + W_{t-1}^\top \mathbf g'_t \big)$  \hfill // \eqref{lip}
    \State Update $(K_t, W_t, \Sigma_t)$ via RLS \eqref{eq:rls_gain}-\eqref{eq:rls_cov}
    \State \textbf{If} {$\|\bar{\boldsymbol{\epsilon}}_t\|_2 > \delta_{\text{thresh}}$} \textbf{Then} $\Sigma_t \leftarrow \sigma_0 I$  \hfill // \ref{reset}
    \State $\hat{\mathbf r}^{adapted}_{t|t} \leftarrow \hat{\mathbf r}^{base}_{t|t} + W_t^\top \mathbf g'_t$ \hfill // \eqref{r_adapt}
    \State Compute $s(t)$ via \eqref{eq:sliding_var}
    \State Update $\hat{\sigma}(t)$ via \eqref{eq:adaptive_law}
    \State Apply control $\tau(t)$ via \eqref{eq:control_input}
\EndWhile
\end{algorithmic}
\end{algorithm}

\vspace{-3mm}
\section{Experimental Validation}
\label{sec:experiments}
\vspace{-1mm}
We evaluate the proposed framework under realistic uncertainties to examine:
(i) predictive fidelity and closed-loop tracking performance,
(ii) the impact of latent-space adaptation via LRA during dynamic variations, and
(iii) the structural contribution of each architectural component through ablation analysis.

\subsection{Experimental Setup}
\label{subsec:setup}

\subsubsection{Hardware Platform}
For experimental validation, we built an aerial manipulator comprising a Tarot-650 quadrotor base (SunnySky V4006 motors, 14-inch propellers) and a 2-DOF planar serial-link manipulator (link lengths $\approx 18$\,cm) actuated by Dynamixel XM430-W210-T motors. The full system weighs $\approx 3.0$\,kg. The quadrotor is equipped with a CUAV X7+ flight controller running customized PX4 firmware, while high-level FDT inference and Adaptive Control (AC) execution happens on an onboard NVIDIA Jetson Orin Nano (8GB). Sensor data from OptiTrack motion capture system (at 120 fps) and IMU were used to measure the necessary pose (position, attitude), velocity and acceleration feedback of the drone. For the manipulator, the joint angular position and velocity are measured by the Dynamixel motors, while the accelerations are computed numerically. Communication leverages Micro XRCE-DDS for low-latency uORB topic exchange, and the manipulator is controlled via the \textit{ros2\_control} Joint Trajectory Controller. The resulting body moments and collective thrust commands are transmitted to flight controller at 50 Hz. 

\subsubsection{Dataset Collection}
To collect diverse training data, we employ a baseline PID controller to execute randomized trajectories that heavily excite the coupled dynamics of the platform and manipulator. Data are gathered under three payload conditions (0\,g, 200\,g, 400\,g) to generate configuration- and load-dependent inertial variations. For each condition, 5 minutes of data are recorded at 100\,Hz, forming the raw time-indexed dataset $\mathcal{D}_{\text{raw}} = \{(\chi^{(i)}, \dot{\chi}^{(i)}, \ddot{\chi}^{(i)},\boldsymbol{\tau}^{(i)})\}_{i=1}^{N_{raw}}$.

\subsubsection{Implementation Details}

The FDT employs $N_{\text{layers}}=2$ encoder layers with latent dimension $d_{\text{model}}=512$ and $n_{\text{heads}}=8$. The decoder is a 2-layer MLP with hidden width $d_{ff}=2048$, projecting the latent representation before a final linear mapping to the residual dimension. The short- and long-horizon windows are set to $T_s=5$ and $T_l=120$ timesteps, respectively. The model performs $k$-step-ahead residual prediction ($k=6$) and is trained using the multi-step residual loss defined in~(\ref{eq:training_loss}) with the Adam optimizer (learning rate $10^{-3}$, batch size $256$) for $100$ epochs with validation-based early stopping.
Online adaptation via LRA employs RLS with forgetting factor $\lambda=0.99$ and initial covariance $\Sigma_0=I$. Covariance reset uses $\delta_{\text{thresh}}=3.2$, $\sigma_0=1$, and exponential smoothing factor $\alpha=0.9$.
The gain parameters for the proposed Adaptive Controller used in all experiments are listed in Table~\ref{tab:controller_params}, where `diag' denotes diagonal matrix. 
The desired roll and pitch of the quadrotor are calculated via \cite{mellinger2011minimum}. 
For all learned models, the input features consist of the current state $\chi_t$, its velocity $\dot{\chi}_t$, and the previous control input $\tau_{t-1}$, since $\tau_t$ is not available at prediction time~\cite{shi2019neural}.

 \begin{table}[!h]
\centering
\caption{Controller Parameter Settings}
\footnotesize
\renewcommand{\arraystretch}{1.2}
\begin{tabular}{l}
\toprule
$\Phi=\text{diag} \lbrace 1.0,\,1.0,\,1.5,\,1.1,\,1.1,\,1.0,\,1.2,\,1.2 \rbrace$ \\
$\Lambda=\text{diag} \lbrace 2.0,\,2.0,\,3.5,\,1.5,\,1.5,\,1.2,\,3.0,\,3.0 \rbrace$ \\
$\bar{M}=\text{diag} \lbrace 2,\,2,\,2,\,0.02,\,0.02,\,0.02,\,0.05,\,0.05 \rbrace$ \\
$\hat{\sigma}(0) = 0.1$, $\nu = 2.0$ \\
\bottomrule
\end{tabular}
\label{tab:controller_params}
\end{table}

\subsubsection{Evaluation Scenarios}
\label{subsubsec:evaluation}

\begin{figure}[h]
\centering
\includegraphics[width=0.98\linewidth]{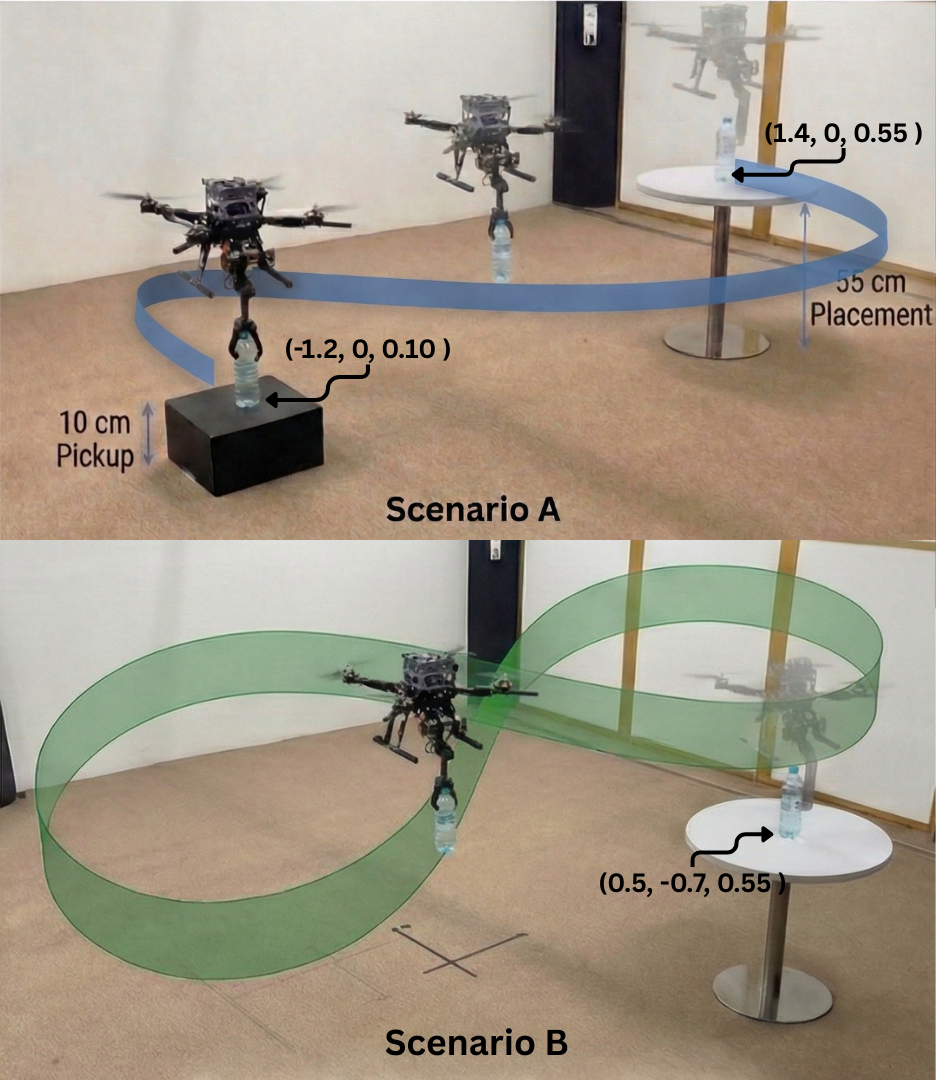}
        \caption{{Dynamic evaluation scenarios.} \textbf{(Top) Scenario A:} An altitude-varying S-shaped pick-and-place maneuver. \textbf{(Bottom) Scenario B:} A continuous Figure-8 trajectory.Unit of coordinates in the figure is in \textit{meters}.}
    \label{fig:model_validate}
\end{figure}

To rigorously assess the framework across varying dynamic regimes, we define a centralized set of physical testing scenarios (cf. Fig. \ref{fig:model_validate}). The evaluation comprises two distinct trajectories: \textit{Scenario A} executes an altitude-varying S-shaped pick-and-place maneuver, grasping an object from a 10\,cm high box and placing it onto a 55\,cm high table, which heavily excites vertical regime shifts and ground-effect transients. \textit{Scenario B} entails a continuous Figure-8 trajectory, picking and placing the object strictly on the 55\,cm table maintaining same height, designed to induce continuous roll-pitch.

Joints of the manipulator move in the range (\ang{-20.1}, \ang{37.2}) and (\ang{65.3}, \ang{77.2}) respectively during the pick-and-place maneuver. 

\begin{table}[h]
\centering
\caption{Model prediction performance comparison}
\label{tab:prediction_results}
\begin{tabular}{@{}lccc@{}}
\toprule
\textbf{Method} & \textbf{RMSE} $\downarrow$ & $\mathbf{R^2}$ $\uparrow$ & \textbf{Inf. (ms)} \\
\midrule
SysID \cite{kaiser2018sparse} & 1.363 & 0.207 & 0.01 \\
GP \cite{torrente2021data} & 0.84 & 0.493 & 17.20 \\
LSTM \cite{rao2024learning} & 0.76 & 0.576 & 1.20 \\
Transformer \cite{vaswani2017attention} & 0.67& 0.623 & 4.10 \\
 Neural-fly \cite{o2022neural} & 0.48 & 0.761 & 9.50 \\
MLP ({\footnotesize last layer}) \cite{saviolo2023active} & 0.41 & 0.802 & 4.50 \\
\midrule
FDT (Base) & 0.29 & 0.912 & 1.86 \\
\textbf{FDT + LRA (Ours)} & \textbf{0.21} & \textbf{0.972} & \textbf{3.02} \\
\bottomrule
\end{tabular}
\end{table}

\vspace{-2mm}
\subsection{Model Validation: Open-Loop Prediction Accuracy}
\label{subsec:model_validation}

For predictive accuracy evaluation, the quadrotor tracks a smooth reference trajectory using a vanilla PID controller for 5 min with a payload of $300$ g is attached to end-effector while $\mathbf r$ is computed at each timestep from \eqref{eq:residual_dynamics}. Performance is quantified via Root Mean Squared Error (RMSE) in residual predictions alongside the Coefficient of Determination ($R^2$)
to evaluate variance capture and record the mean inference time in milliseconds ({Inf. (ms)}), as shown in Table \ref{tab:prediction_results}.

\subsection{Trajectory Tracking \& Disturbance Rejection}
\label{subsec:tracking}

We evaluate trajectory tracking performance, for scenarios A and B and shown in Fig. \ref{fig:model_validate}, by applying the control law~\eqref{eq:adaptive_control_law} and computing the root-mean-square error (RMSE) of error signal $e(t)$ for each method. Table \ref{tab:tracking_results} shows the performance of compared baselines under same testing scenarios.

\begin{table*}[t]
\centering
\caption{\footnotesize Tracking RMSE under In-Distribution (300\,g) and Out-of-Distribution (500\,g) payloads for Scenario A and Scenario B.}
\label{tab:tracking_results}
\vspace{0.15cm}
\resizebox{\textwidth}{!}{
\begin{tabular}{lcccccccc}
\toprule
& \multicolumn{4}{c}{\textbf{Scenario A}} & \multicolumn{4}{c}{\textbf{Scenario B}} \\
\cmidrule(lr){2-5} \cmidrule(lr){6-9}
\multirow{2}{*}{Method}
& \multicolumn{2}{c}{In-Distribution (300\,g)} 
& \multicolumn{2}{c}{Out-of-Distribution (500\,g)}
& \multicolumn{2}{c}{In-Distribution (300\,g)} 
& \multicolumn{2}{c}{Out-of-Distribution (500\,g)} \\
\cmidrule(lr){2-3} \cmidrule(lr){4-5} \cmidrule(lr){6-7} \cmidrule(lr){8-9}
& Slow (0.5\,m/s) $\downarrow$
& Fast (1.0\,m/s) $\downarrow$
& Slow (0.5\,m/s) $\downarrow$
& Fast (1.0\,m/s) $\downarrow$
& Slow (0.5\,m/s) $\downarrow$
& Fast (1.0\,m/s) $\downarrow$
& Slow (0.5\,m/s) $\downarrow$
& Fast (1.0\,m/s) $\downarrow$ \\
\midrule
\multicolumn{9}{l}{\textit{Baseline}} \\
$\mathcal{L}_1$ Adaptive \cite{wu20251}
& 1.222 & 1.560 & 1.711 & 2.223
& 0.981 & 1.202 & 1.293 & 1.617 \\
\midrule
\multicolumn{9}{l}{\textit{Offline learning Baselines}} \\
GP \cite{torrente2021data}
& 0.752 & 0.960 & 1.053 & 1.368
& 0.604 & 0.740 & 0.796 & 0.995 \\
LSTM \cite{rao2024learning}
& 0.680 & 0.868 & 0.952 & 1.237
& 0.546 & 0.669 & 0.720 & 0.900 \\
GPT-2 \cite{chen2021decision}
& 0.599 & 0.765 & 0.839 & 1.090
& 0.481 & 0.589 & 0.634 & 0.793 \\
\midrule
\multicolumn{9}{l}{\textit{Online adaptation Baselines}} \\
Neural-fly \cite{o2022neural}
& 0.429 & 0.547 & 0.600 & 0.780
& 0.344 & 0.422 & 0.454 & 0.567 \\
MLP ({\footnotesize last layer}) \cite{saviolo2023active}
& 0.366 & 0.467 & 0.512 & 0.665
& 0.294 & 0.360 & 0.387 & 0.484 \\
\midrule
\multicolumn{9}{l}{\textbf{Ours}} \\
\textbf{FDT (Ours)}
& 0.258 & 0.330 & 0.362 & 0.470
& 0.207 & 0.254 & 0.274 & 0.342 \\
\textbf{FDT+LRA (Ours)}
& \textbf{0.215} & \textbf{0.275} & \textbf{0.301} & \textbf{0.391}
& \textbf{0.173} & \textbf{0.212} & \textbf{0.228} & \textbf{0.285} \\
\bottomrule
\end{tabular}
}
\end{table*}

\vspace{-1mm}

\subsection{Analysis}
\label{subsec:analysis}

As evidenced by Tables~\ref{tab:prediction_results} and \ref{tab:tracking_results}, across both prediction and tracking metrics, performance correlates with how explicitly each method models structured residual dynamics and adapts to time-varying system dynamics. 
Sparse identification methods are constrained by predefined functional libraries, leading to systematic bias under strongly coupled inertial and aerodynamic effects. 
GP models capture smooth disturbances but struggle with scalability and high-dimensional coupling. Sequence models such as LSTM and Transformer architectures mitigate temporal compounding error; however, recurrent memory or time-tokenized attention alone entangles instantaneous dynamics with slow regime evolution, limiting structural interpretability and generalization.
While $\mathcal{L}_1$-adaptive control ensures robustness under bounded matched uncertainties, it lacks an explicit residual model, causing performance degradation under structured, configuration-dependent multiscale dynamics. Adaptive model learning approaches improve the model performance, but their mechanisms differ fundamentally. Meta-learned models such as Neural-Fly rely on interpolation within trained task families and lack explicit structural decomposition of residual components, restricting extrapolation under abrupt inertial changes. 
Gradient-based last-layer updates adapt parameters in weight space, coupling representation and identification and resulting in slower, optimizer-sensitive convergence. 
In contrast, our framework constrains adaptation to a linear latent subspace with closed-form updates, yielding well-conditioned, rapid regime adaptation while preserving structured residual representation. Notably, FDT+LRA preserves millisecond-scale inference, indicating that structured factorization and latent adaptation impose minimal computational overhead in real-time operation.

\subsection{Ablation Studies}
\label{subsec:ablation}

\begin{table}[t]
\renewcommand{\arraystretch}{1.3}
\centering
\caption{Ablation Study on Test Set}
\label{tab:ablation}
\scalebox{0.85}{
\begin{tabular}{@{}p{5.2cm}cc@{}}
\toprule
\textbf{Variant} & \textbf{RMSE} $\downarrow$ & $\Delta$ \textbf{RMSE vs. Full(\%)} \\
\midrule
\textbf{FDT + LRA (Ours)} & \textbf{0.283} & -- \\
\midrule
w/o Online LRA (\ref{subsubsec:_online_lra}) & 0.37 & +30.7\% \\
w/o Global Token (\textbf{g}) & 0.692 & +144.5\% \\
w/o Short-Horizon Context Encoding (\ref{subsubsec:context_stream}) & 0.778 & +174.9\% \\
w/o Long-Horizon Memory (\ref{subsubsec:memory_stream}) & 0.782 & +176.3\% \\
\bottomrule
\end{tabular}
}
\end{table}

To validate the architectural design, we perform ablation studies on our model, in Scenario B under a 500 g payload and 1.0 m/s mean velocity, we use model prediction RMSE to compare the degraded model with proposed full model (percentage $\Delta$ {RMSE vs. Full}) quantifying the impact on predictive accuracy, shown in Table~\ref{tab:ablation}.

The ablation results confirm that structured multi-scale factorization is critical for accurate residual modeling. Removing the long-horizon memory or short-horizon context encoding leads to the largest degradation, indicating that both slow regime evolution and instantaneous cross-variable coupling are indispensable for capturing AAM residual dynamics under payload variation. Eliminating the global token significantly impairs performance, demonstrating that explicit cross-variable aggregation is essential for coherent latent regime representation. Finally, disabling the online LRA produces a moderate but consistent degradation, verifying that latent-space adaptation enhances robustness under regime shifts while relying on the structural backbone of the full FDT.

\vspace{-1mm}
\section{Conclusion and Future Work}
\label{sec:conclusion}

We introduced a predictive–adaptive framework for compensating nonstationary residual dynamics in aerial manipulators. The proposed Factorized Dynamics Transformer structurally separates cross-variable inertial coupling from long-horizon aerodynamic memory, enabling accurate multi-scale residual prediction. A Latent Residual Adapter performs lightweight online linear identification in latent space, allowing rapid recovery from unseen payload shifts while preserving real-time feasibility.
Real-world experiments demonstrate improved prediction accuracy, faster disturbance rejection, and superior closed-loop tracking compared to classical and learning-based baselines, underscoring the value of structured factorization and continuous latent adaptation for robust aerial manipulation.

The proposed factorization and latent linear adaptation  assume that regime shifts remain locally linearizable in the learned latent representation, which may break down under highly abrupt or extreme dynamic transitions.
Future work will incorporate uncertainty-aware residual modeling and explicit transient regime-change detection to improve adaptation beyond the initial training envelope.
\vspace{-1mm}
\section{DECLARATION OF AI-ASSISTANCE}
ChatGPT 5.2 was used only for minor language refinement; all technical content was created and verified by the authors, who take full responsibility for the publication.

\bibliographystyle{IEEEtran}
\bibliography{root}

@ARTICLE{ollero2022past,
  author={Ollero, Anibal and Tognon, Marco and Suarez, Alejandro and Lee, Dongjun and Franchi, Antonio},
  journal={IEEE Transactions on Robotics}, 
  title={Past, Present, and Future of Aerial Robotic Manipulators}, 
  year={2022},
  volume={38},
  number={1},
  pages={626-645},
  doi={10.1109/TRO.2021.3084395}
}

@inproceedings{gahlawat2020l1,
  title={L1-GP: L1 adaptive control with Bayesian learning},
  author={Gahlawat, Aditya and Zhao, Pan and Patterson, Andrew and Hovakimyan, Naira and Theodorou, Evangelos},
  booktitle={Learning for dynamics and control},
  pages={826--837},
  year={2020},
  organization={PMLR}
}

@inproceedings{shi2019neural,
  title={Neural lander: Stable drone landing control using learned dynamics},
  author={Shi, Guanya and Shi, Xichen and O'Connell, Michael and Yu, Rose and Azizzadenesheli, Kamyar and Anandkumar, Animashree and Yue, Yisong and Chung, Soon-Jo},
  booktitle={2019 International Conference on Robotics and Automation (ICRA)},
  pages={9784--9790},
  year={2019},
  organization={IEEE}
}

@inproceedings{richards2021adaptive,
  title={Adaptive-Control-Oriented Meta-Learning for Nonlinear Systems},
  author={Richards, SM and Azizan, N and Slotine, J-JE and Pavone, M},
  booktitle={Robotics: Science and Systems},
  year={2021}
}

@article{mohajerin2019multistep,
  title={Multistep prediction of dynamic systems with recurrent neural networks},
  author={Mohajerin, Nima and Waslander, Steven L},
  journal={IEEE Transactions on Neural Networks and Learning Systems},
  volume={30},
  number={11},
  pages={3370--3383},
  year={2019},
  publisher={IEEE}
}

@inproceedings{das2025dronediffusion,
  title={Dronediffusion: Robust quadrotor dynamics learning with diffusion models},
  author={Das, Avirup and Yadav, Rishabh Dev and Sun, Sihao and Sun, Mingfei and Kaski, Samuel and Pan, Wei},
  booktitle={2025 IEEE International Conference on Robotics and Automation (ICRA)},
  pages={1604--1610},
  year={2025},
  organization={IEEE}
}

@ARTICLE{yadav2025integrated,
  author={Yadav, Rishabh Dev and Jones, Brycen and Gupta, Saksham and Sharma, Amitabh and Sun, Jiefeng and Zhao, Jianguo and Roy, Spandan},
  journal={IEEE/ASME Transactions on Mechatronics}, 
  title={An Integrated Approach to Aerial Grasping: Combining a Bistable Gripper With Adaptive Control}, 
  year={2025},
  pages={1-12},
  doi={10.1109/TMECH.2025.3586888}
}

@article{o2022neural,
  title={Neural-fly enables rapid learning for agile flight in strong winds},
  author={O’Connell, Michael and Shi, Guanya and Shi, Xichen and Azizzadenesheli, Kamyar and Anandkumar, Anima and Yue, Yisong and Chung, Soon-Jo},
  journal={Science Robotics},
  volume={7},
  number={66},
  pages={eabm6597},
  year={2022},
  publisher={American Association for the Advancement of Science}
}

@ARTICLE{saviolo2023active,
  author={Saviolo, Alessandro and Frey, Jonathan and Rathod, Abhishek and Diehl, Moritz and Loianno, Giuseppe},
  journal={IEEE Transactions on Robotics}, 
  title={Active Learning of Discrete-Time Dynamics for Uncertainty-Aware Model Predictive Control}, 
  year={2024},
  volume={40},
  pages={1273-1291},
  doi={10.1109/TRO.2023.3339543}
}

@inproceedings{jiahao2023online,
  title={Online dynamics learning for predictive control with an application to aerial robots},
  author={Jiahao, Tom Z and Chee, Kong Yao and Hsieh, M Ani},
  booktitle={Conference on Robot Learning},
  pages={2251--2261},
  year={2023},
  organization={PMLR}
}

@article{lew2022safe,
  title={Safe active dynamics learning and control: A sequential exploration--exploitation framework},
  author={Lew, Thomas and Sharma, Apoorva and Harrison, James and Bylard, Andrew and Pavone, Marco},
  journal={IEEE Transactions on Robotics},
  volume={38},
  number={5},
  pages={2888--2907},
  year={2022},
  publisher={IEEE}
}

@article{kaiser2018sparse,
  title={Sparse identification of nonlinear dynamics for model predictive control in the low-data limit},
  author={Kaiser, Eurika and Kutz, J Nathan and Brunton, Steven L},
  journal={Proceedings of the Royal Society A},
  volume={474},
  number={2219},
  pages={20180335},
  year={2018},
  publisher={The Royal Society Publishing}
}

@article{torrente2021data,
  title={Data-driven mpc for quadrotors},
  author={Torrente, Guillem and Kaufmann, Elia and F{\"o}hn, Philipp and Scaramuzza, Davide},
  journal={IEEE Robotics and Automation Letters},
  volume={6},
  number={2},
  pages={3769--3776},
  year={2021},
  publisher={IEEE}
}

@article{chen2021decision,
  title={Decision transformer: Reinforcement learning via sequence modeling},
  author={Chen, Lili and Lu, Kevin and Rajeswaran, Aravind and Lee, Kimin and Grover, Aditya and Laskin, Misha and Abbeel, Pieter and Srinivas, Aravind and Mordatch, Igor},
  journal={Advances in Neural Information Processing Systems},
  volume={34},
  pages={15084--15097},
  year={2021}
}

@inproceedings{mellinger2011minimum,
  title={Minimum snap trajectory generation and control for quadrotors},
  author={Mellinger, Daniel and Kumar, Vijay},
  booktitle={2011 IEEE International Conference on Robotics and Automation (ICRA)},
  pages={2520--2525},
  year={2011},
  organization={IEEE}
}

@article{meng2020survey,
  title={Survey on aerial manipulator: System, modeling, and control},
  author={Meng, Xiangdong and He, Yuqing and Han, Jianda},
  journal={Robotica},
  volume={38},
  number={7},
  pages={1288--1317},
  year={2020},
  publisher={Cambridge University Press}
}

@inproceedings{cao2024computation,
  author    = {Wenhan Cao and Alexandre Capone and Rishabh Yadav and Sandra Hirche and Wei Pan},
  title     = {Computation-Aware Learning for Stable Control with Gaussian Process},
  booktitle = {Proceedings of Robotics: Science and Systems (RSS)},
  year      = {2024}
}

@article{saviolo2022physics,
  title={Physics-inspired temporal learning of quadrotor dynamics for accurate model predictive trajectory tracking},
  author={Saviolo, Alessandro and Li, Guanrui and Loianno, Giuseppe},
  journal={IEEE Robotics and Automation Letters},
  volume={7},
  number={4},
  pages={10256--10263},
  year={2022},
  publisher={IEEE}
}

@article{orsag2017dexterous,
  title={Dexterous aerial robots—mobile manipulation using unmanned aerial systems},
  author={Orsag, Matko and Korpela, Christopher and Bogdan, Stjepan and Oh, Paul},
  journal={IEEE Transactions on Robotics},
  volume={33},
  number={6},
  pages={1453--1466},
  year={2017},
  publisher={IEEE}
}

@article{ruggiero2018aerial,
  title={Aerial manipulation: A literature review},
  author={Ruggiero, Fabio and Lippiello, Vincenzo and Ollero, Anibal},
  journal={IEEE Robotics and Automation Letters},
  volume={3},
  number={3},
  pages={1957--1964},
  year={2018},
  publisher={IEEE}
}

@article{saviolo2023learning,
  title={Learning quadrotor dynamics for precise, safe, and agile flight control},
  author={Saviolo, Alessandro and Loianno, Giuseppe},
  journal={Annual Reviews in Control},
  volume={55},
  pages={45--60},
  year={2023},
  publisher={Elsevier}
}

@inproceedings{rao2024learning,
  title={Learning long-horizon predictions for quadrotor dynamics},
  author={Rao, Pratyaksh Prabhav and Saviolo, Alessandro and Ferrari, Tommaso Castiglione and Loianno, Giuseppe},
  booktitle={2024 IEEE/RSJ International Conference on Intelligent Robots and Systems (IROS)},
  pages={12758--12765},
  year={2024},
  organization={IEEE}
}

@article{wu20251,
  title={L 1 Quad: L 1 Adaptive Augmentation of Geometric Control for Agile Quadrotors With Performance Guarantees},
  author={Wu, Zhuohuan and Cheng, Sheng and Zhao, Pan and Gahlawat, Aditya and Ackerman, Kasey A and Lakshmanan, Arun and Yang, Chengyu and Yu, Jiahao and Hovakimyan, Naira},
  journal={IEEE Transactions on Control Systems Technology},
  volume={33},
  number={2},
  pages={597--612},
  year={2025},
  publisher={IEEE}
}

@article{vaswani2017attention,
  title={Attention is all you need},
  author={Vaswani, Ashish and Shazeer, Noam and Parmar, Niki and Uszkoreit, Jakob and Jones, Llion and Gomez, Aidan N and Kaiser, {\L}ukasz and Polosukhin, Illia},
  journal={Advances in neural information processing systems},
  volume={30},
  year={2017}
}

@inproceedings{sharma2025impedance,
  title={Impedance and Stability Targeted Adaptation for Aerial Manipulator with Unknown Coupling Dynamics},
  author={Sharma, Amitabh and Gupta, Saksham and Singh, Shivansh Pratap and Yadav, Rishabh Dev and Song, Hongyu and Pan, Wei and Roy, Spandan and Baldi, Simone},
  booktitle={2025 25th International Conference on Control, Automation and Systems (ICCAS)},
  pages={471--476},
  year={2025},
  organization={IEEE}
}

@article{yadav2024modular,
  title={Modular adaptive aerial manipulation under unknown dynamic coupling forces},
  author={Yadav, Rishabh Dev and Dantu, Swati and Pan, Wei and Sun, Sihao and Roy, Spandan and Baldi, Simone},
  journal={IEEE/ASME Transactions on Mechatronics},
  volume={30},
  number={4},
  pages={2688--2698},
  year={2024},
  publisher={IEEE}
}

@article{ujjawal2025aermani,
  title={AERMANI-Diffusion: Regime-Conditioned Diffusion for Dynamics Learning in Aerial Manipulators},
  author={Ujjawal, Samaksh and Singh, Shivansh Pratap and Nair, Naveen Sudheer and Yadav, Rishabh Dev and Pan, Wei and Roy, Spandan},
  journal={arXiv preprint arXiv:2512.10773},
  year={2025}
}

@article{yadav2026physics,
  title={Physics-aware sparse learning and selective online adaptation for euler-lagrange robot dynamics},
  author={Yadav, Rishabh Dev and Ujjawal, Samaksh and Sun, Sihao and Roy, Spandan and Pan, Wei},
  journal={arXiv preprint arXiv:2606.09640},
  year={2026}
}

@article{yadav2026learning,
  title={Learning Cross-Coupled and Regime Dependent Dynamics for Aerial Manipulation},
  author={Yadav, Rishabh Dev and Ujjawal, Samaksh and Sun, Sihao and Roy, Spandan and Pan, Wei},
  journal={arXiv preprint arXiv:2605.14805},
  year={2026}
}

@article{yadav2025arcade,
  title={ARCADE: Adaptive Robot Control with Online Changepoint-Aware Bayesian Dynamics Learning},
  author={Yadav, Rishabh Dev and Das, Avirup and Song, Hongyu and Kaski, Samuel and Pan, Wei},
  journal={arXiv preprint arXiv:2512.14331},
  year={2025}
}

@article{mishra2025aermani,
  title={AERMANI-VLM: Structured Prompting and Reasoning for Aerial Manipulation with Vision Language Models},
  author={Mishra, Sarthak and Yadav, Rishabh Dev and Das, Avirup and Gupta, Saksham and Pan, Wei and Roy, Spandan},
  journal={arXiv preprint arXiv:2511.01472},
  year={2025}
}

@article{singh2026aerograb,
  title={AeroGrab: A Unified Framework for Aerial Grasping in Cluttered Environments},
  author={Singh, Shivansh Pratap and Nair, Naveen Sudheer and Ujjawal, Samaksh and Mishra, Sarthak and Patil, Soham and Yadav, Rishabh Dev and Roy, Spandan},
  journal={arXiv preprint arXiv:2603.15097},
  year={2026}
}

@article{mishra2026aeroplace,
  title={AeroPlace-Flow: Language-Grounded Object Placement for Aerial Manipulators via Visual Foresight and Object Flow},
  author={Mishra, Sarthak and Yadav, Rishabh Dev and Nair, Naveen and Pan, Wei and Roy, Spandan},
  journal={arXiv preprint arXiv:2603.07744},
  year={2026}
}

@article{song2025soranav,
  title={SoraNav: Adaptive UAV Task-Centric Navigation via Zeroshot VLM Reasoning},
  author={Song, Hongyu and Yadav, Rishabh Dev and Guo, Cheng and Pan, Wei},
  journal={arXiv preprint arXiv:2510.25191},
  year={2025}
}
                
\end{document}